# Universal Observatory Graphs for Distributed Sky Coverage and Artificial Intelligence-Based Interplanetary Routing

**Mohammed Abdel Razek**

## Abstract

Distributed space observatories could mitigate the geometric limitations of individual telescopes, but their collective visibility and communication performance require a common mathematical representation. Here, we propose the Universal Observatory Graph (UOG), an AI-driven framework for distributed astronomical observation across the Solar System. The proposed architecture models autonomous observatories located at the Sun–planet $L_2$ Lagrange points as nodes in a weighted graph, while communication links are represented as graph edges characterized by multi-objective physical and operational metrics, including interplanetary distance, communication latency, transmission power, and link reliability. The resulting graph provides a unified mathematical representation of a cooperative interplanetary observatory network. This proposal examines a six-observatory Solar System configuration comprising Earth, Mars, Jupiter, Saturn, Uranus and Neptune. Instantaneous sky coverage is evaluated independently using a 200,000-direction Fibonacci sphere, a 2,000,000-direction fixed-seed Monte Carlo calculation and deterministic spherical integration. All three methods yield complete network union coverage, approximately 0.43% complete six-observatory intersection and approximately 24.96% mean pairwise Jaccard similarity under the adopted pointing model. Communication routing is subsequently formulated as a finite-horizon Markov decision process and solved using tabular Q-learning. The reward balances node participation and a distance-dependent reliability proxy against distance, light-time latency and a distance-squared transmission-power proxy. The learned Earth–Saturn–Uranus–Neptune route is also the highest-discounted-return route among all 41 feasible simple paths under the four-hop constraint. The framework provides a reproducible baseline for sequential coverage assessment and multi-objective routing; time-dependent ephemerides, mission-specific visibility, calibrated link budgets and scalable graph policies remain future work.

## Introduction

Space observatories avoid atmospheric absorption and turbulence and have therefore transformed infrared and multi-wavelength astronomy. The James Webb Space Telescope (JWST), operating in a halo orbit around the Sun–Earth $L_2$ region, illustrates the scientific value of a thermally controlled deep-space platform. Its commissioning demonstrated observatory and instrument performance that meets or exceeds many pre-launch expectations[1,2].

Thermal protection also constrains instantaneous observability. JWST maintains its telescope and instruments behind a sunshield and is restricted to solar elongations of approximately 85°–135°[3]. Targets near the ecliptic are therefore available only during finite visibility windows. This limitation is not a defect of a particular instrument; it follows from the geometry and thermal requirements of a single sun shielded observing platform. The growth of time-domain and multi-messenger astronomy makes such temporal restrictions increasingly consequential. Low-latency gravitational-wave alerts enable rapid electromagnetic follow-up[4] while tidal disruption events and other transients evolve over observationally important timescales[5] A distributed set of observatories with complementary fields of regard could reduce instantaneous blind regions and provide redundant access to selected parts of the sky.

Deep-space communication introduces a second constraint. Interplanetary links are characterized by long propagation delays, distance-dependent link budgets and potential interruptions. Delay/disruption-tolerant networking was developed for precisely such challenged environments[6], and distributed spacecraft autonomy is an active area of mission research[7] These developments motivate a representation that combines observatory geometry with a weighted communication network. Here we introduce the UOG as a general graph-theoretic representation of distributed observatories. We distinguish two computational layers. The first quantifies instantaneous scientific coverage through deterministic Fibonacci sampling, stochastic Monte Carlo validation and spherical integration. The second optimizes communication routing over the fixed weighted graph using tabular Q-learning. This separation is deliberate: the current code evaluates coverage before routing but does not place visibility masks or observation windows inside the learning environment.

We apply the framework to six hypothetical observatories associated with the Sun–planet $L_2$ regions of Earth, Mars, Jupiter, Saturn, Uranus and Neptune. The case study tests whether spatially distributed pointing axes can provide complementary instantaneous coverage and whether a graph-based reinforcement-learning agent can select multi-hop communication routes under an explicitly defined multi-objective reward.

## Results

### A six-node Universal Observatory Graph

The Solar System demonstration contains six observatory nodes and 15 undirected links, forming a complete graph. Each node represents a hypothetical observatory associated with a Sun–planet $L_2$ region; the implementation does not simulate a resolved halo orbit. Node pointing axes are determined from prescribed heliocentric longitudes, whereas communication distances are supplied by a fixed symmetric distance matrix.

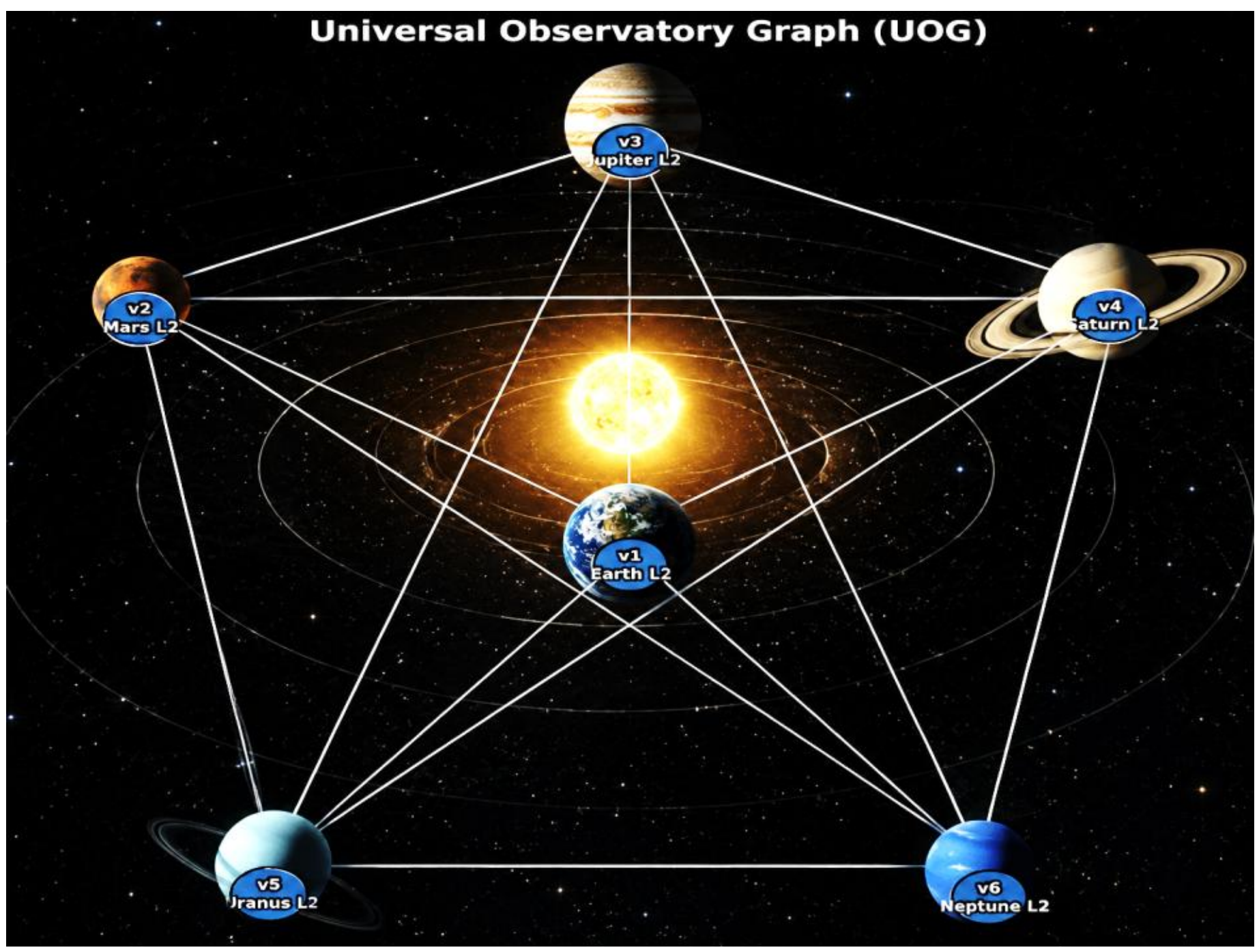

**Figure 1 | Complete six-node Universal Observatory Graph. The nodes correspond to observatories associated with Earth, Mars, Jupiter, Saturn, Uranus and Neptune. All 15 undirected links are available to the routing environment; the layout is schematic and is not a spatially scaled representation of the Solar System.**

## Individual coverage expands towards the outer Solar System

The adopted upper elongation limit increases from 135° at Earth to values approaching 150° at large heliocentric distance. Consequently, deterministic spherical integration gives individual instantaneous coverage values ranging from 39.713% for Earth to 47.440% for Neptune. The Fibonacci and Monte Carlo estimates reproduce the same monotonic trend.

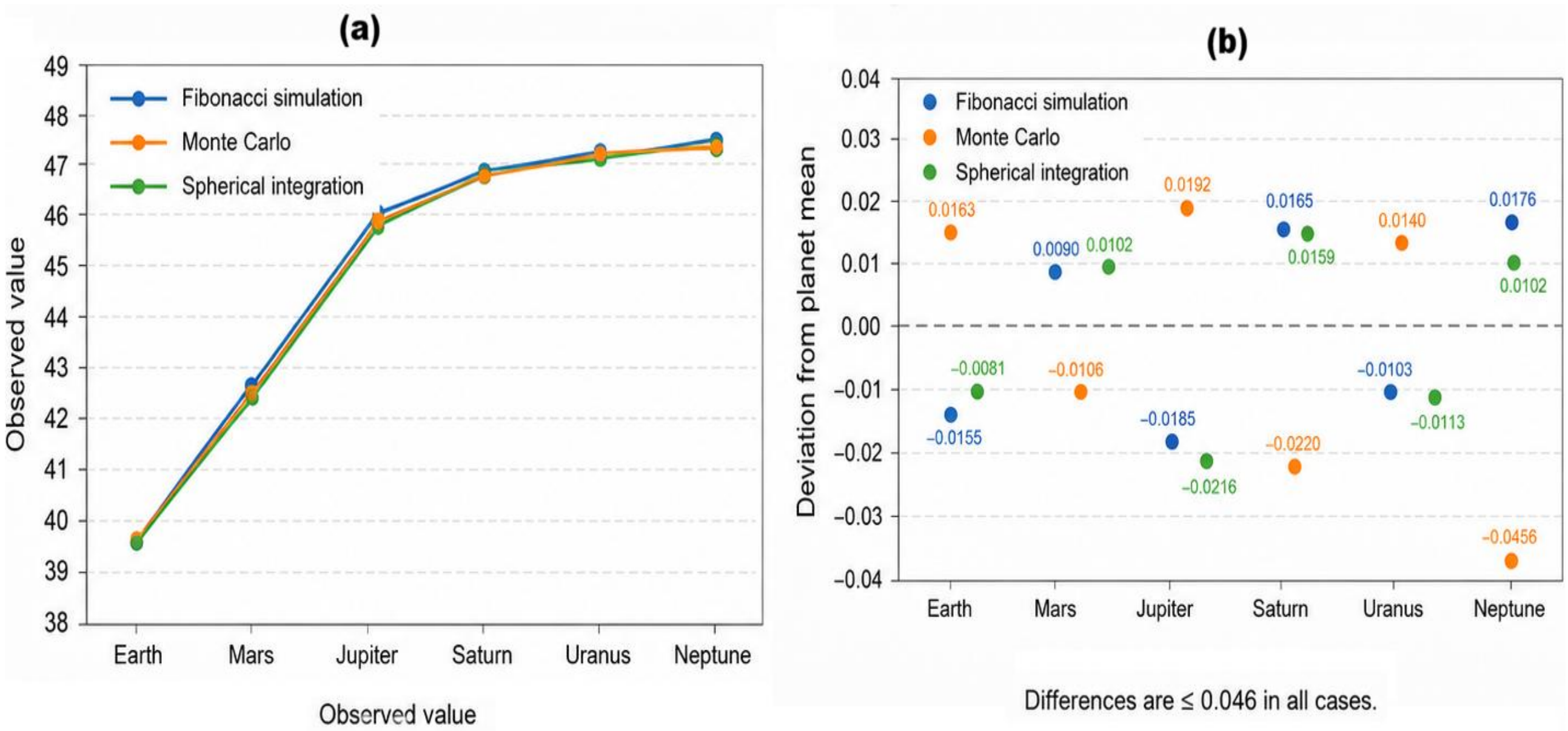

**Figure 2 | Individual observatory coverage across validation methods. (a) Line-chart comparison: Panel (a) shows the values obtained using the Fibonacci simulation, Monte Carlo, and spherical integration methods for Earth, Mars, Jupiter, Saturn, Uranus, and Neptune. All three curves follow almost identical increasing trends, from approximately 39.7 for Earth to about 47.4 for Neptune, indicating very close agreement among the three computational approaches. (d) Deviation analysis: Panel (d) illustrates the deviation of each computational result from the mean value calculated for the corresponding planet. The deviations remain close to zero for all three methods, demonstrating a high level of numerical consistency. The largest differences are still very small, confirming that the Fibonacci simulation, Monte Carlo, and spherical integration approaches produce highly comparable results across all six planets.**

## Complementary fields of regard produce full network union

All three methods achieve complete network union coverage under the adopted geometry. The complete six-observatory intersection is much smaller: 0.4280% for the Fibonacci discretization, 0.43055% for Monte Carlo and 0.428406% for spherical integration. The network therefore reaches full union coverage through complementary viewing regions rather than through a large region shared by every observatory. Figure 3 presents a dual-axis comparison of network-level coverage and redundancy metrics across three computational methods: Fibonacci simulation, Monte Carlo, and spherical integration. The diverging bars, referenced to the left y-axis, represent deviations in complete intersection percentage from the overall mean. Monte Carlo shows a small positive deviation, whereas the Fibonacci and spherical integration methods exhibit slight negative deviations, indicating only minor differences among the three approaches. The black line with circular markers, referenced to the right y-axis, shows the mean pairwise Jaccard percentage. The values remain very close, ranging from approximately 24.959% to 24.962%, with Fibonacci simulation producing the highest value, Monte Carlo the lowest, and spherical integration lying between them. Overall, the figure demonstrates strong agreement among the three computational methods, with only negligible variation in both complete intersection and mean pairwise Jaccard metrics.

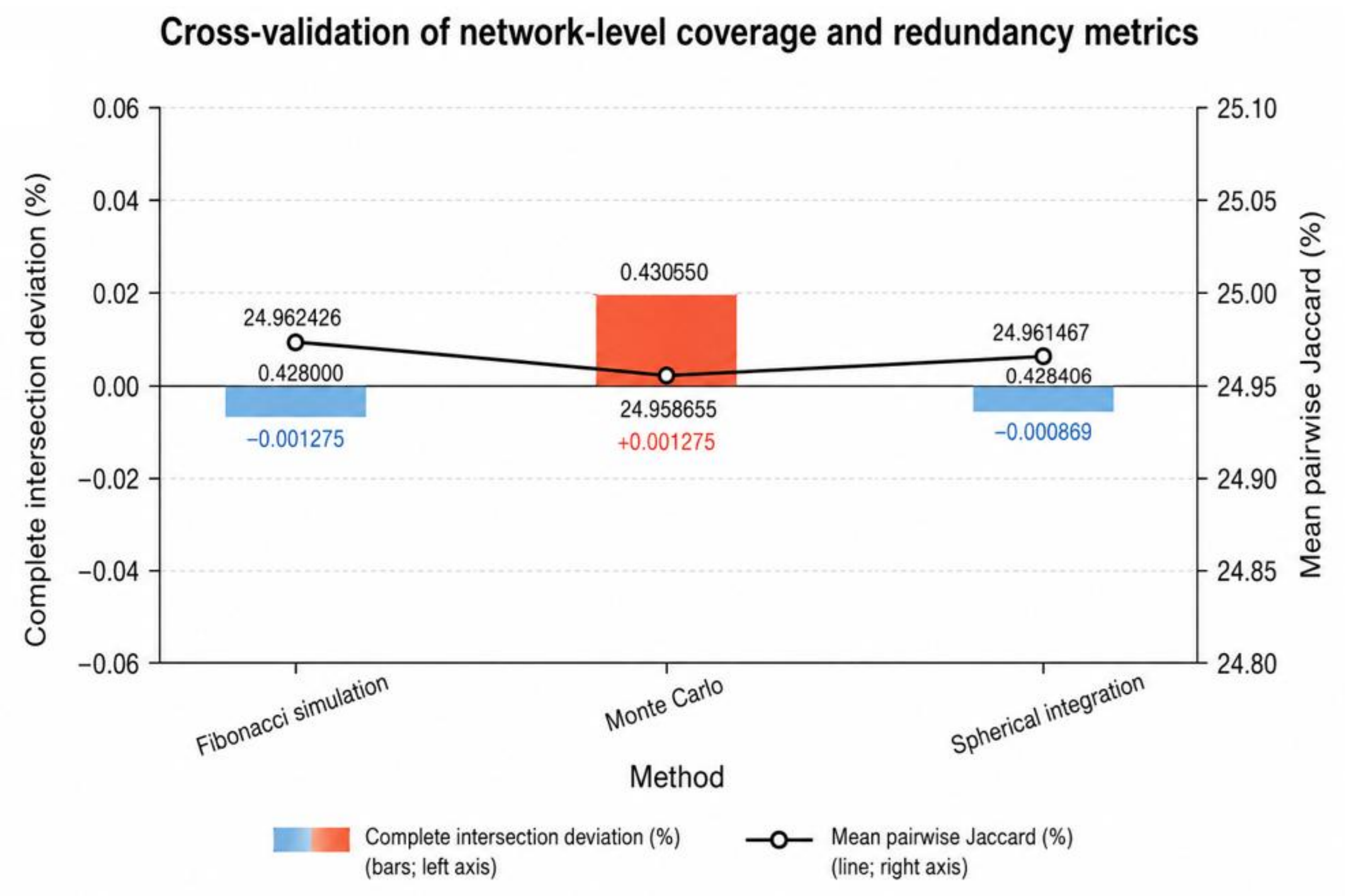

**Figure 3 | Cross-validation of network-level coverage and redundancy metrics. Diverging bars show deviations in complete-intersection coverage from the mean across the Fibonacci simulation, Monte Carlo and spherical integration methods (left axis), whereas the black line with open-circle markers shows the corresponding mean pairwise Jaccard values (right axis). All three approaches yield closely matched results, with only minor deviations in complete-intersection coverage and negligible variation in pairwise Jaccard similarity, indicating strong numerical consistency across the independent computational methods.**

## Pairwise overlap reveals moderate average redundancy

The mean pairwise Jaccard similarity is 24.9624%, 24.9587% and 24.9615% for the Fibonacci, Monte Carlo and spherical-integration calculations, respectively. Individual pairs span a broader range. Spherical integration gives the lowest Jaccard similarity for Mars–Uranus (11.770%) and the highest for Earth–Mars (38.216%). The network therefore combines substantial overlaps for some neighboring viewing geometries with strong complementarity for others.

Figure 4 presents a pairwise sky overlap matrix for six observatories associated with Earth, Mars, Jupiter, Saturn, Uranus and Neptune. Each matrix cell reports the percentage of sky coverage shared between a pair of observatories, with the colour scale indicating the magnitude of the overlap. Higher values are represented by lighter yellow tones, whereas lower overlaps are shown in purple to blue shades.

The diagonal elements correspond to the individual sky coverage of each observatory and contain the largest values, increasing from 39.7% for Earth to approximately 47.4% for Neptune. Off-diagonal values quantify the overlap between different observatory pairs and are substantially lower, generally ranging from about 9.5% to 24.3%. The strongest pairwise overlap occurs between Jupiter and Saturn (24.3%), while the smallest overlap is observed between Mars and Uranus (9.5%). Overall, the matrix demonstrates that the six observatories provide complementary sky coverage with limited redundancy between most pairs.

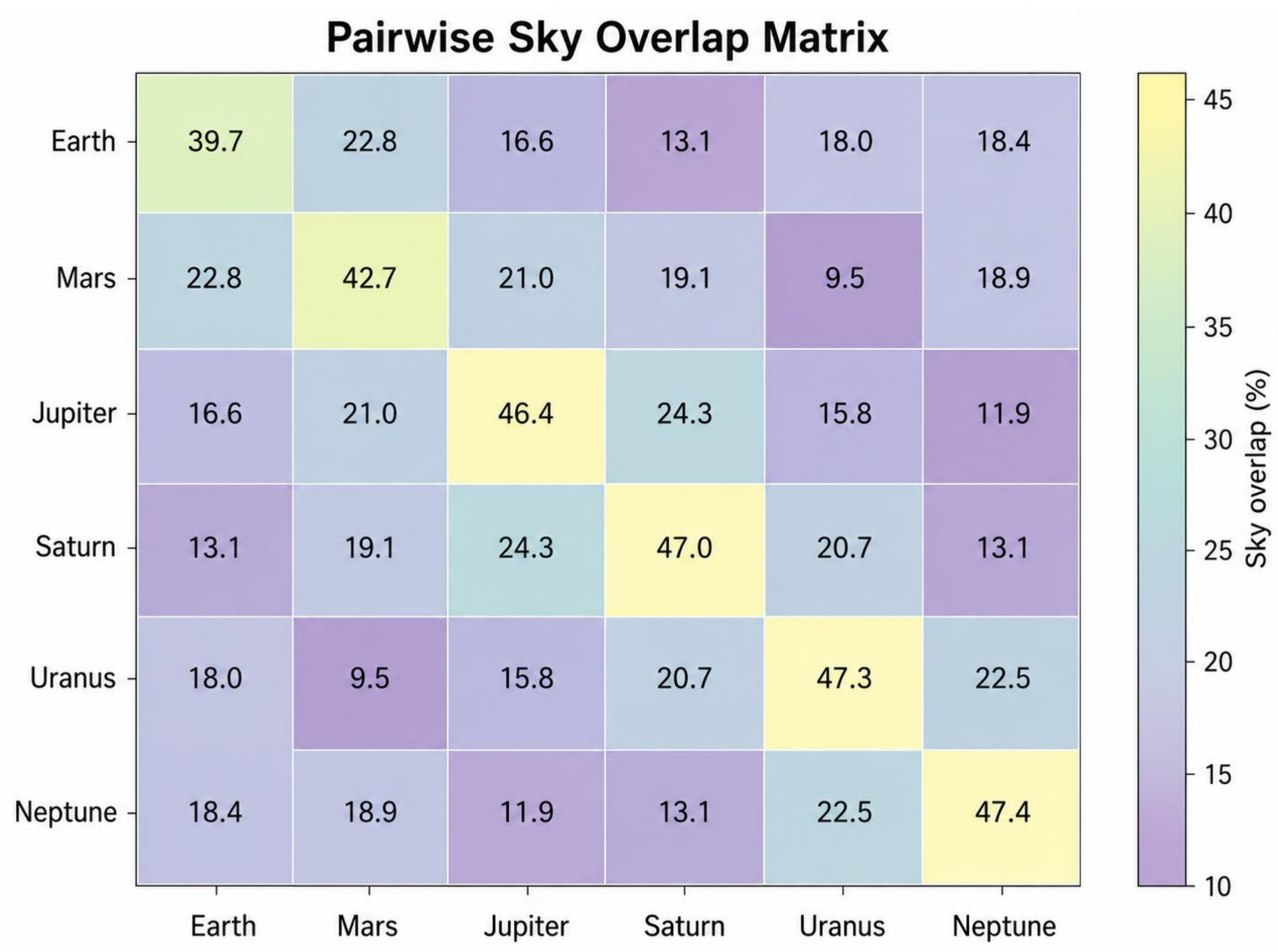

**Figure 4 | Pairwise sky-overlap matrix for the Fibonacci-sphere calculation. Diagonal values represent individual coverage; off-diagonal values quantify the fraction of the celestial sphere shared by each observatory pair.**

## Communication weights vary systematically with distance

The prescribed pairwise distances range from 0.524 au for Earth–Mars to 29.110 au for Earth–Neptune. Because latency is proportional to distance, the corresponding one-way delays range from 261.5 s to 14,525.9 s. The normalized power proxy grows quadratically, whereas the reliability proxy decreases exponentially. These quantities should be interpreted as comparative edge attributes: they are not ephemeris-resolved distances, complete radio-frequency link budgets or measured failure probabilities.

## Tabular Q-learning selects a multi-hop route

The archived visualization records a representative learned route from Earth to Neptune through Uranus. This route contains three nodes, has a cumulative distance of 29.110 au, a one-way latency of 14,525.89 s, a cumulative power proxy of 450.531 and an end-to-end reliability proxy of 0.05442. Its terminal reward under the implemented objective is 26.904.

This result must be interpreted as a representative stochastic training outcome. The routing code initializes exploration through NumPy's global random-number generator but does not fix a reinforcement-learning seed.

Repeated training may therefore return different greedy routes. Moreover, the current experiment does not compare the learned route systematically with an exhaustive optimum or a shortest-path baseline. The result demonstrates operation of the routing framework, not proof of globally optimal routing.

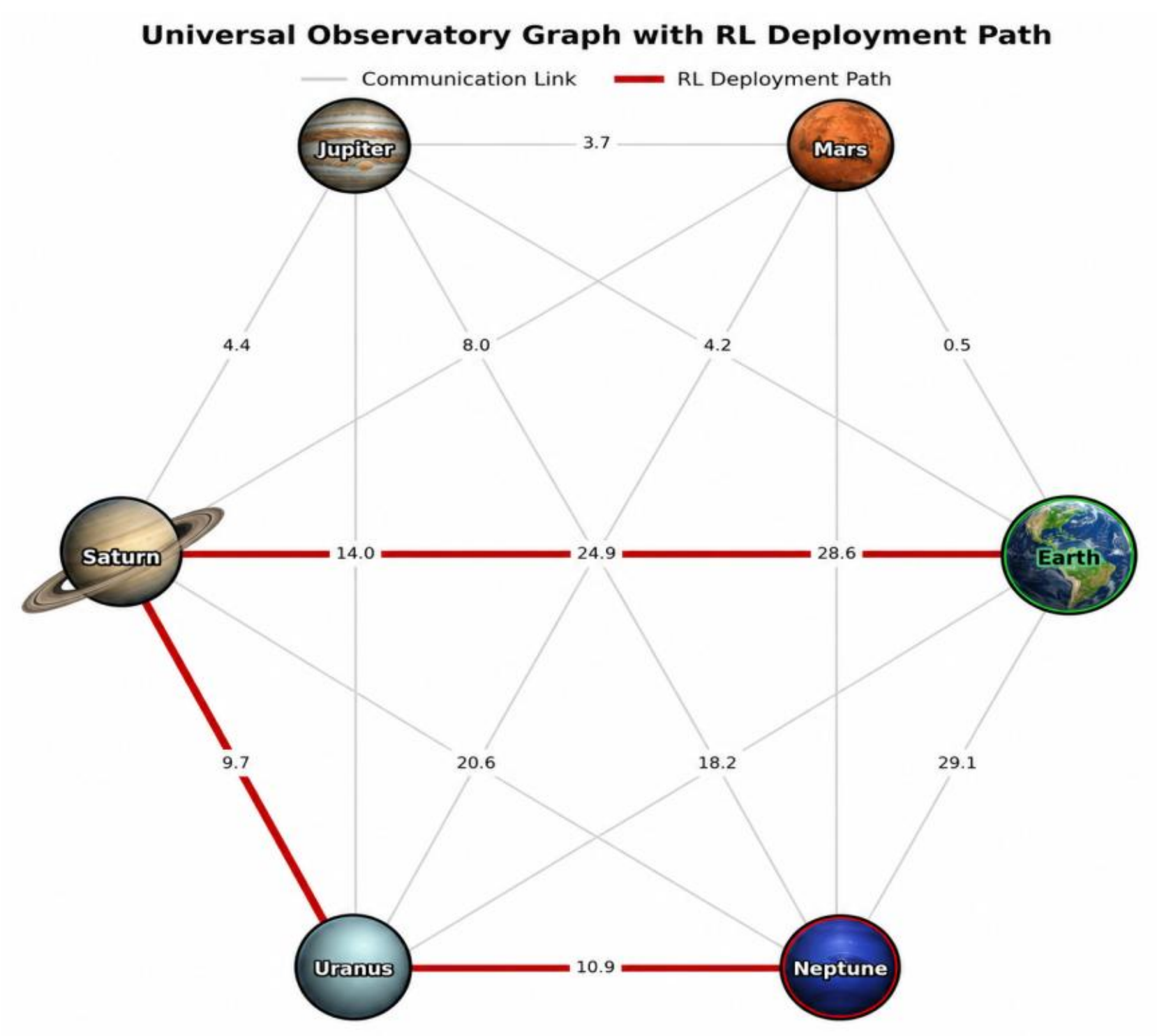

**Figure 5 | Distance-dependent communication attributes of the UOG. The plotted values are derived from the fixed distance matrix and the model relations for latency, power proxy and reliability proxy.**

## Discussion

The principal scientific result is geometric: distributed observatories with different Sun-relative axes can collectively remove the instantaneous blind regions of an individual sun shielded telescope. Under the adopted elongation model, no single observatory accesses more than 47.45% of the sky, yet the union of six visibility regions reaches the complete sphere. The small common intersection shows that this gain is dominated by complementarity rather than universal redundancy.

The agreement among the three numerical approaches strengthens the internal consistency of this conclusion. Fibonacci lattices are well suited to spherical area estimation[9] Monte Carlo directions provide an independent stochastic calculation[10] and direct spherical integration supplies a reference for the idealized annular geometry[11]. Their agreement does not, however, validate the physical pointing model itself. The adopted distance-dependent upper elongation limit is a study-specific hypothesis and requires mission-level thermal and attitude analysis.

The communication model provides a compact way to explore competing routing considerations. Light-time latency follows directly from distance, while the distance-squared power term retains the dominant scaling implied by free-space propagation. The exponential reliability term is phenomenological. Its principal role is to introduce a preference against long individual links; it should not be interpreted as a calibrated failure probability.

The reward does not seek the shortest path alone. By rewarding the number of visited nodes, it favours participation by multiple observatories, whereas distance, latency and power penalize inefficient detours. This scalarization makes the routing objective transparent, but the selected coefficients determine the preferred trade-off. A future study should therefore include reward-weight sensitivity analysis and comparisons with exact enumeration, Dijkstra-type baselines and multi-objective Pareto methods.

### Limitations

Several limitations define the scope of the present results. First, observatory positions are constructed from prescribed mean orbital radii and angular locations, while the communication-distance matrix is independently specified. The analysis does not use time-resolved planetary ephemerides or propagate halo or Lissajous orbits. Second, visibility is instantaneous and axisymmetric; planetary occultation, roll-dependent restrictions, stray light, thermal history, scheduling and detector constraints are omitted.

Third, the communication graph is complete and static. Link availability, antenna pointing, solar conjunction, bandwidth, data volume and disruption-tolerant store-and-forward behaviour are not simulated. Fourth, sky coverage does not constrain the reinforcement-learning environment. The model neither constructs a target-dependent feasible node set nor jointly optimizes deployment and topology.

Finally, the algorithm is tabular Q-learning despite the graph-based environment. The graph-embedding and policy-network functions in the supplied source are placeholders and do not implement a graph neural network. The absence of a fixed training seed also limits route-level reproducibility. These restrictions motivate a future formulation that couples time-dependent visibility to graph states, fixes and reports all random seeds, and benchmarks learned policies against exact and deterministic routing algorithms.

## Conclusion

We introduced the Universal Observatory Graph as a common representation of scientific visibility and communication relationships in a distributed planetary observatory network. For a six-node Solar System demonstration, three independent methods consistently produced complete instantaneous union coverage under the adopted pointing geometry, while approximately 0.43% complete intersection and 24.96% mean pairwise Jaccard similarity showed that the result is driven by complementary fields of regard.

Communication routing over the fixed weighted graph was formulated as a finite-horizon Markov decision process and addressed using tabular Q-learning. The objective balances node participation and a reliability proxy against distance, latency and a power proxy. The framework therefore extends beyond single-metric shortest-path routing, but the archived route should be interpreted as a representative stochastic outcome rather than proof of global optimality.

The present work establishes a reproducible baseline in which sky coverage and communication routing are evaluated sequentially. Time-dependent ephemerides, mission-specific pointing models, realistic link budgets, target-aware state constraints and scalable graph-neural-network policies provide the next steps towards a physically operational distributed observatory architecture.

# Methods

### Universal Observatory Graph

To establish a general and scalable framework for autonomous distributed astronomical observatories, we introduce the Universal Observatory Graph (UOG), a graph-theoretic representation that models candidate observatory locations and their communication relationships independently of any specific planetary system. Unlike conventional mission architectures that optimize individual observatories or isolated planetary systems, the UOG provides a unified mathematical framework applicable to any stellar system for which orbital parameters are available. Consequently, the proposed framework is not restricted to the Solar System but is designed to support future interplanetary and extrasolar observatory networks as astronomical exploration expands beyond the Solar System.

The Universal Observatory Graph is formally represented by the weighted graph

$$\mathrm{UOG} = G(V, E, W),$$

where $V$ denotes the set of candidate observatory nodes, $E$ represents the set of feasible communication or interplanetary transfer links between observatories, and $W$ defines the associated edge-weight function. Each graph node corresponds to a dynamically stable orbital location suitable for long-duration astronomical observations, while each edge represents a physically feasible connection characterized by engineering and operational constraints. Each candidate observatory node is defined as

$$v_i = (S_k, P_j, L_2),$$

where $S_k$ denotes the host star, $P_j$ represents the $j^{\mathrm{th}}$ planet orbiting the host star, and $L_2$ denotes the corresponding star–planet $L_2$ Lagrange point selected as the deployment location of the astronomical observatory. This formulation allows the graph to scale naturally with the number of planets within a stellar system while remaining independent of the characteristics of any planetary system.

The choice of the $L_2$ Lagrange point is motivated by its unique suitability for astronomical observatories. Among the five classical Lagrange points, $L_2$ provides a favorable operational environment by enabling continuous anti-stellar pointing while maintaining stable thermal conditions through permanent shielding from the host star. This geometry minimizes stray-light contamination, simplifies spacecraft attitude control, and supports the cryogenic operating conditions required by large infrared observatories. The successful operation of missions such as the James Webb Space Telescope at the Sun–Earth $L_2$ point demonstrates the scientific and engineering advantages of this orbital environment. Employing the $L_2$ point for every planet further establishes a homogeneous set of observatory nodes sharing comparable operational characteristics, thereby simplifying network design and enabling systematic optimization across multiple planetary systems. A communication or transfer link between two observatories is represented by an edge $e_{ij} \in E$, which exists whenever a direct communication channel or an interplanetary transfer trajectory between nodes $v_i$ and $v_j$ is physically feasible. Each edge is associated with a multidimensional weight vector $w_{ij} = \left(d_{ij}, \tau_{ij}, P_{ij}, C_{ij}, R_{ij}\right)$, where $d_{ij}$ is the inter-node distance, $\tau_{ij}$ is the communication latency, $P_{ij}$ is the required transmission power, $C_{ij}$ represents the deployment and operational cost, and $R_{ij}$ denotes the communication reliability. Unlike conventional shortest-path formulations that minimize a single scalar cost, this multidimensional representation enables simultaneous consideration of scientific performance, engineering feasibility, and operational efficiency. The resulting graph forms a distributed observatory network in which every node contributes both observational capability and communication functionality. Because every observatory may simultaneously serve as a scientific instrument and an interplanetary relay station, the proposed architecture naturally supports multi-hop communication, autonomous data routing, and fault-tolerant network operation. Such capabilities become increasingly important as observatory networks expand across large interplanetary distances where direct communication with Earth may become impractical or inefficient.

The UOG is intentionally formulated independently of any planetary system. Consequently, it can represent observatory networks spanning the Solar System, extrasolar planetary systems, or future multi-system astronomical infrastructures without modification of the underlying mathematical model. In the present work, the Solar System is selected as the first validation case because its planetary ephemerides, orbital dynamics, and engineering constraints are accurately characterized, enabling quantitative evaluation of the proposed framework under realistic conditions. The same methodology can subsequently be extended to any stellar system for which sufficient orbital information becomes available. Figure 6 illustrates the conceptual structure of the Universal Observatory Graph. Each planetary $L_2$ point constitutes a graph node, while feasible communication or transfer links establish the network connectivity that is subsequently optimized by the proposed artificial intelligence framework.

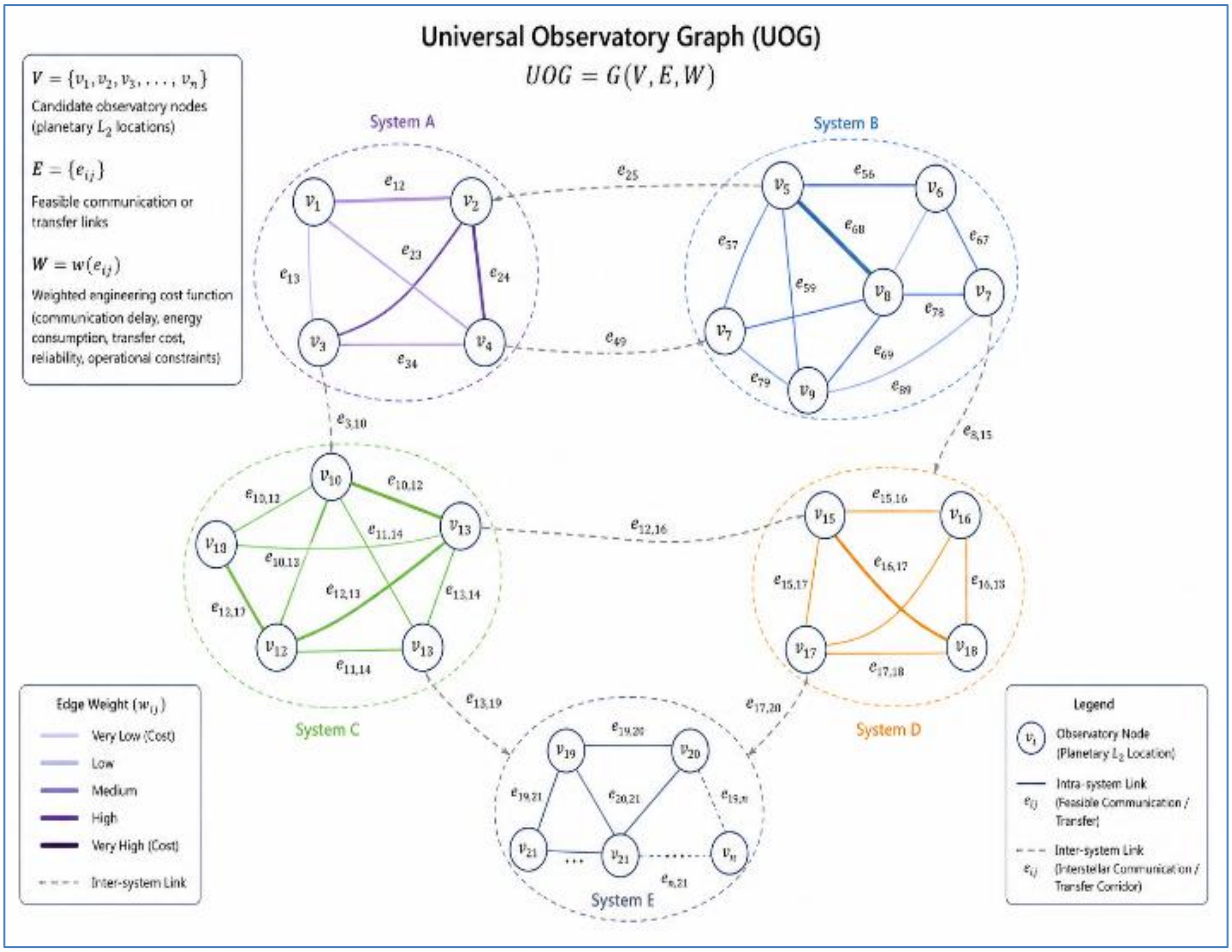

**Figure 6 | Conceptual representation of the Universal Observatory Graph, $UOG = G(V, E, W)$, a generalized weighted graph describing distributed space observatories across arbitrary stellar systems. Nodes ($v_i$) correspond to candidate observatories at planetary $L_2$ locations, whereas edges ($e_{ij}$) represent feasible communication or transfer links weighted by engineering and operational costs. The resulting graph constitutes the mathematical foundation of the proposed artificial intelligence framework for optimizing large-scale observatory network architectures**

**Universal Observatory Graph for Solar System**

The observatory network is represented as a weighted undirected graph, consistent with standard network notation[8],

$$G = (V, E, W), \tag{1}$$

where $V$ is the observatory set, $E$ is the communication-link set and $W$ assigns a multidimensional attribute vector to each edge. The Solar System implementation uses six nodes and includes every unordered node pair, giving

$$|E| = \frac{n(n-1)}{2} = \frac{6(6-1)}{2} = 15.$$

**Observatory geometry**

Each observatory is assigned a heliocentric position from a prescribed orbital radius $r_i$ and longitude $\theta_i$. The unit vector from observatory $i$ towards the Sun is $\hat{s}_i$. A candidate viewing direction $\omega$ lies on the unit celestial sphere, and its solar elongation is

$$\alpha_i(\omega) = \cos^{-1}(\hat{s}_i \cdot \omega). \tag{2}$$

The instantaneous visibility region is the spherical annulus

$$S_i = \{\omega \in S^2 : \alpha_{min,i} \leq \alpha_i(\omega) \leq \alpha_{max,i}\}. \tag{3}$$

The minimum elongation is 85°. The maximum elongation is prescribed as

$$\alpha_{\max,i} = 135^\circ + 15^\circ \left(1 - \frac{1}{r_i}\right), \tag{4}$$

with $r_i$ in astronomical units. This expression is introduced for the present case study and is not derived from a spacecraft thermal model. The Earth value matches the operational JWST elongation range documented by STScI[3].

**Coverage metrics**

Individual coverage is the normalized solid angle of $S_i$,

$$C_i = \frac{A(S_i)}{4\pi}. \tag{5}$$

For an axisymmetric annulus, direct integration gives

$$C_i = \frac{\cos(\alpha_{\min,i}) - \cos(\alpha_{\max,i})}{2}. \tag{6}$$

Network union and complete intersection are

$$C_{\text{union}} = \frac{A\left(\bigcup_i S_i\right)}{4\pi}, \; C_{\text{intersection}} = \frac{A\left(\bigcap_i S_i\right)}{4\pi}. \tag{7}$$

Pairwise redundancy is measured using the Jaccard coefficient[18]

$$J_{ij} = \frac{A(S_i \cap S_j)}{A(S_i \cup S_j)}, \tag{8}$$

and the mean pairwise value is

$$\bar{J}_{\text{pair}} = \frac{1}{\binom{n}{2}} \sum_{i<j} J_{ij}. \tag{9}$$

**Fibonacci-sphere simulation**

The primary calculation discretizes the sphere using $N_{\text{F}} = 200{,}000$approximately equal-area Fibonacci directions. Fibonacci lattices reduce spherical area-estimation error relative to latitude–longitude sampling.[9] For $k = 0, \dots, N_{\text{F}} - 1$, the code constructs

$$y_k = 1 - \frac{2k}{N_{\text{F}} - 1}, \ \phi_k = k\pi(3 - \sqrt{5}), \tag{10}$$

and

$$\omega_k = \left(\sqrt{1 - y_k^2}, \cos\phi_k \, , \ y_k \sqrt{1 - y_k^2} \sin\phi_k\right). \tag{11}$$

A binary mask $M_{ik}$records whether direction $\omega_k$satisfies equation (3). Individual coverage is the mean of $M_{ik}$, while union and complete intersection are obtained through Boolean OR and AND operations across observatory masks.

**Monte Carlo simulation**

The independent stochastic calculation uses $N_{\text{MC}} = 2{,}000{,}000$directions and random seed 42. Uniform solid-angle sampling is obtained from $z \sim U(-,11)$and $\phi \sim U(0, 2\pi)$,[10] with

$$\omega = \left(\sqrt{1 - z^2}, \cos\phi \, , \ \sqrt{1 - z^2} \sin\phi \ \ z\right). \tag{12}$$

Coverage is estimated from the fraction of samples satisfying each Boolean visibility condition. For an estimated fraction $\hat{p}$, the binomial sampling standard error is

$$\text{SE}(\hat{p}) = \sqrt{\frac{\hat{p}(1 - \hat{p})}{N_{\text{MC}}}}. \tag{13}$$

**Spherical integration**

For a Boolean region $B(\theta, \phi)$, the reference solid angle is

$$A(B) = \int_0^{2\pi} \int_0^{\pi} \mathbb{I}\,[B(\theta, \phi)] \sin\theta \ \, \mathrm{d}\theta \, \mathrm{d}\phi. \tag{14}$$

Pairwise annular intersections are reduced to a one-dimensional quadrature over the admissible azimuthal interval $\Delta\phi_{ij}(\theta)$, following the geometry of intersecting spherical caps[11]

$$A(S_i \cap S_j) = \int \Delta\,\phi_{ij}(\theta)\sin\theta\;\mathrm{d}\theta. \tag{15}$$

The complete network union and intersection are evaluated by direct integration of their Boolean indicator functions.

**Communication-edge model**

Each edge is assigned

$$w_{ij} = (D_{ij}, L_{ij}, P_{ij}, R_{ij}). \tag{16}$$

The code uses a fixed pairwise distance matrix $D$ in astronomical units. The astronomical unit is exactly $149{,}597{,}870{,}700$m[12] yielding a one-way light time of approximately 499.0048 s. The implementation uses the rounded relation

$$L_{ij} = 499D_{ij}. \tag{17}$$

The Friis free-space relation implies an inverse-square dependence of received power on distance under fixed antenna and wavelength assumptions [13]. The code therefore adopts the relative power proxy

$$P_{ij} = D_{ij}^2. \tag{18}$$

The distance-dependent reliability proxy is

$$R_{ij} = \exp\left(-\frac{D_{ij}}{10}\right). \tag{19}$$

Although the exponential form resembles a constant-hazard reliability function[14] distance is substituted for exposure time and the 10-au scale is study-specific. $R_{ij}$is therefore a normalized preference term.

**Markov decision process**

Routing is represented as a finite-horizon Markov decision process[15,16],

$$\mathcal{M} = (\mathcal{S}, \mathcal{A}, P, \mathcal{R}, \gamma). \tag{20}$$

At time $t$, the state is $s_t = (v_t, \mathcal{V}_t)$,

where $v_t$is the current node and $\mathcal{V}_t$is the visited-node set. Available actions are unvisited neighbours,

$$\mathcal{A}(s_t) = \{v_j : (v_t, v_j) \in E \text{ and } v_j \notin \mathcal{V}_t\}. \tag{21}$$

An action moves the agent deterministically to $v_j$and updates the visited set. An episode terminates at Neptune, when the path reaches the six-node limit or when no unvisited neighbour remains.

**Reward and Q-learning**

For a successful route $\rho$, cumulative distance, latency and power are additive, whereas reliability is the product of the link proxies. The terminal reward is

$$\mathcal{R}_\rho = 15N_\rho + 20R_{\rho,\mathrm{link}} - 0.10D_\rho - 0.0005L_\rho - 0.02P_\rho \quad (22)$$

Here $N_\rho$ is the number of distinct visited nodes. Failure to reach the goal produces a reward of $-100$. All coefficients are study-specific scalarization weights.

Q-values are updated using the standard off-policy rule[17],

$$Q(s_t, a_t) \leftarrow Q(s_t, a_t) + \alpha \left[r_t + \gamma \max_{a'} Q(s_{t+1}, a') - Q(s_t, a_t)\right] \quad (23)$$

Training uses 5,000 episodes, learning rate $\alpha = 0.10$, discount factor $\gamma = 0.95$, initial exploration $\varepsilon = 1.0$, minimum exploration $\varepsilon_{\min} = 0.05$and multiplicative decay 0.995. Actions are chosen $\varepsilon$-greedily during training and greedily after training. The reinforcement-learning random seed is not fixed in the current code.

**Computational complexity**

Neighbour and edge retrieval scan the edge list and therefore require $O(m)$time for $m$links. A loop-free episode contains at most $n - 1$transitions, giving a conservative $O(nm)$time bound per episode. For a complete graph, $m = O(n^2)$, so the worst-case bound becomes $O(n^3)$. Because the state includes the visited-node subset, the theoretical tabular state space is $O(n2^n)$. This scaling is tractable for six nodes but motivates function approximation for larger networks.

**Software and reproducibility**

The supplied implementation is written in Python and uses NumPy for numerical operations, Matplotlib for visualization and NetworkX for graph figures. Coverage tables were read directly from the archived CSV outputs. The Monte Carlo seed is fixed at 42. The routing experiment uses 5,000 training episodes but does not set a random seed; route-level results should therefore be regenerated with an explicit seed before formal publication.

## Acknowledgements

The author acknowledges the developers and maintainers of the open scientific Python ecosystem used in this study. No specific funding information was supplied for this manuscript.

## Author contributions

M.A.R. conceived the Universal Observatory Graph framework, developed the methodology and software, performed the numerical analysis, interpreted the results, prepared the visualizations and wrote the manuscript.

## Competing interests

The author declares no competing interests.

## Supplementary information

The following tables reproduce the principal numerical outputs contained in the UOG-V04 archive. Percentages are reported relative to the complete celestial sphere.

**Supplementary Table S1 | Pairwise intersection coverage obtained using each method.**

| Pair | Fibonacci simulation | Monte Carlo | Spherical integration |
|---|---|---|---|
| Earth - Mars | 22.793500 | 22.800400 | 22.800140 |
| Earth - Jupiter | 16.622500 | 16.626150 | 16.621206 |
| Earth - Saturn | 13.148000 | 13.164950 | 13.147314 |
| Earth - Uranus | 18.039500 | 18.062950 | 18.045878 |
| Earth - Neptune | 18.351000 | 18.357250 | 18.351593 |
| Mars - Jupiter | 20.965500 | 20.965100 | 20.960777 |
| Mars - Saturn | 19.120500 | 19.117900 | 19.114266 |
| Mars - Uranus | 9.484000 | 9.445900 | 9.483944 |
| Mars - Neptune | 18.939000 | 18.914250 | 18.942709 |
| Jupiter - Saturn | 24.277500 | 24.275950 | 24.273413 |
| Jupiter - Uranus | 15.803500 | 15.830800 | 15.803318 |
| Jupiter - Neptune | 11.860000 | 11.865850 | 11.856031 |
| Saturn - Uranus | 20.740000 | 20.750850 | 20.738283 |
| Saturn - Neptune | 13.145000 | 13.074500 | 13.147314 |

| Pair | Fibonacci simulation | Monte Carlo | Spherical integration |
|---|---|---|---|
| Uranus - Neptune | 22.453000 | 22.419900 | 22.445284 |

**Supplementary Table S2 | Pairwise union coverage and Jaccard similarity. All numerical entries are percentages.**

| Pair | Fibonacci union | MC union | Spherical union | Fibonacci J | MC J | Spherical J |
|---|---|---|---|---|---|---|
| Earth - Mars | 59.6590 | 59.6643 | 59.6612 | 38.2063 | 38.2145 | 38.2160 |
| Earth - Jupiter | 69.4330 | 69.4989 | 69.4388 | 23.9403 | 23.9229 | 23.9365 |
| Earth - Saturn | 73.5160 | 73.4923 | 73.5237 | 17.8845 | 17.9134 | 17.8817 |
| Earth - Uranus | 68.9815 | 69.0140 | 68.9817 | 26.1512 | 26.1729 | 26.1604 |
| Earth - Neptune | 68.8020 | 68.7643 | 68.8016 | 26.6722 | 26.6959 | 26.6732 |
| Mars - Jupiter | 68.1315 | 68.1501 | 68.1343 | 30.7721 | 30.7631 | 30.7639 |
| Mars - Saturn | 70.5850 | 70.5296 | 70.5918 | 27.0886 | 27.1062 | 27.0772 |
| Mars - Uranus | 80.5785 | 80.6213 | 80.5787 | 11.7699 | 11.7164 | 11.7698 |
| Mars - Neptune | 71.2555 | 71.1976 | 71.2455 | 26.5790 | 26.5659 | 26.5879 |
| Jupiter - Saturn | 69.0310 | 69.0318 | 69.0314 | 35.1690 | 35.1663 | 35.1629 |
| Jupiter - Uranus | 77.8620 | 77.8967 | 77.8580 | 20.2968 | 20.3228 | 20.2976 |
| Jupiter - Neptune | 81.9375 | 81.9062 | 81.9309 | 14.4744 | 14.4871 | 14.4708 |
| Saturn - Uranus | 73.5340 | 73.5089 | 73.5341 | 28.2046 | 28.2290 | 28.2023 |
| Saturn - Neptune | 81.2610 | 81.2298 | 81.2507 | 16.1763 | 16.0957 | 16.1812 |
| Uranus - Neptune | 72.3100 | 72.3042 | 72.3093 | 31.0510 | 31.0077 | 31.0407 |

## Supplementary algorithm

**Algorithm S1** | Tabular Q-learning over the fixed UOG

1. Initialize the Q-table and ε = 1.0.
2. For each of 5,000 episodes, reset the agent to Earth and initialize the visited set.
3. Identify unvisited neighbours of the current node.
4. Select a random admissible action with probability ε; otherwise select the highest-Q action.
5. Traverse the selected edge and update distance, latency, power and reliability totals.
6. Terminate at Neptune, at the six-node path limit or at a dead end.
7. Assign the terminal multi-objective reward or the −100 failure penalty.
8. Update Q(s,a) using equation (23) and decay ε by 0.995 to a minimum of 0.05.
9. After training, reset to Earth and repeatedly select the highest-Q admissible action.